\documentclass[journal]{IEEEtran}
\usepackage{amsmath,amsfonts}
\usepackage{algorithmic}
\usepackage{array}
\usepackage[caption=false,font=normalsize,labelfont=sf,textfont=sf]{subfig}
\usepackage{multirow}
\usepackage{xcolor}
\usepackage{cite}
\usepackage{textcomp}
\usepackage{stfloats}
\usepackage{url}
\usepackage{verbatim}
\usepackage{graphicx}
\usepackage{booktabs}

\def\BibTeX{{\rm B\kern-.05em{\sc i\kern-.025em b}\kern-.08em
    T\kern-.1667em\lower.7ex\hbox{E}\kern-.125emX}}
\usepackage{balance}

\begin{document}
\title{End-to-End Learnable Multi-Scale Feature Compression for VCM}

\author{Yeongwoong~Kim,
        Hyewon~Jeong,
        Janghyun~Yu,
        Younhee~Kim,
        Jooyoung~Lee,
        Se~Yoon~Jeong,
        and~Hui~Yong~Kim,~\IEEEmembership{Member,~IEEE,}
        
    \thanks{Manuscript received Apr. 14, 2023; revised May. 27, 2023; accepted Jul. 24, 2023. This work was supported by Institute of Information \& Communications Technology Planning \& Evaluation (IITP) grant funded by the Korea government (MSIT) (No. 2020-0-00011, Video Coding for Machine), Institute of Information and Communications Technology Planning and Evaluation (IITP) grant funded by the Korea Government (MSIT) (No. 2021-0-02068, Artificial Intelligence Innovation Hub), and Institute of Information \& communications Technology Planning \& Evaluation (IITP) grant funded by the Korea government(MSIT) (No. RS-2022-00155911, Artificial Intelligence Convergence Innovation Human Resources Development (Kyung Hee University)). (\it{Corresponding author: Prof. Hui Yong Kim})}
    \thanks{Hui Yong Kim, Yeongwoong Kim, Hyewon Jeong and Janghyun Yu are with the Visual Media Lab, School of Computing, Kyung Hee University, Republic of Korea (\textit{e}-mail: \{hykim.v, duddnd7575, woni980911, janghyun1120\}@khu.ac.kr).}
    \thanks{Younhee Kim, Jooyoung Lee and Se Yoon Jeong are with the Electronics and Telecommunications Research Institute (ETRI), Republic of Korea (\textit{e}-mail: \{kimyounhee, leejy1003, seyoonjeong\}@etri.re.kr).
    }
        }

\markboth{IEEE Transactions on Circuits and Systems for Video Technology}%
{}

\makeatletter
\def\ps@IEEEtitlepagestyle{
  \def\@oddfoot{\mycopyrightnotice}
  \def\@evenfoot{}
}
\def\mycopyrightnotice{
  {\footnotesize
  \begin{minipage}{\textwidth}
  \centering
  Copyright~\copyright~2023 IEEE. Personal use of this material is permitted. However, permission to use this material for any other purposes must be obtained from the IEEE by sending an email to pubs-permissions@ieee.org.
  \end{minipage}
  }
}

\maketitle

\begin{abstract}
The proliferation of deep learning-based machine vision applications has given rise to a new type of compression, so called video coding for machine (VCM). VCM differs from traditional video coding in that it is optimized for machine vision performance instead of human visual quality. In the feature compression track of MPEG-VCM, multi-scale features extracted from images are subject to compression. Recent feature compression works have demonstrated that the versatile video coding (VVC) standard-based approach can achieve a BD-rate reduction of up to 96\% against MPEG-VCM feature anchor. However, it is still sub-optimal as VVC was not designed for extracted features but for natural images. Moreover, the high encoding complexity of VVC makes it difficult to design a lightweight encoder without sacrificing performance. 
To address these challenges, we propose a novel multi-scale feature compression method that enables both the end-to-end optimization on the extracted features and the design of lightweight encoders. The proposed model combines a learnable compressor with a multi-scale feature fusion network so that the redundancy in the multi-scale features is effectively removed. Instead of simply cascading the fusion network and the compression network, we integrate the fusion and encoding processes in an interleaved way. Our model first encodes a larger-scale feature to obtain a latent representation and then fuses the latent with a smaller-scale feature. This process is successively performed until the smallest-scale feature is fused and then the encoded latent at the final stage is entropy-coded for transmission. 
The results show that our model outperforms previous approaches by at least 52\% BD-rate reduction and has $\times5$ to $\times27$ times less encoding time for object detection. It is noteworthy that our model can attain near-lossless task performance with only 0.002-0.003\% of the uncompressed feature data size.
\end{abstract}


\begin{figure*}[!t]
\centering
\includegraphics[width=\linewidth]{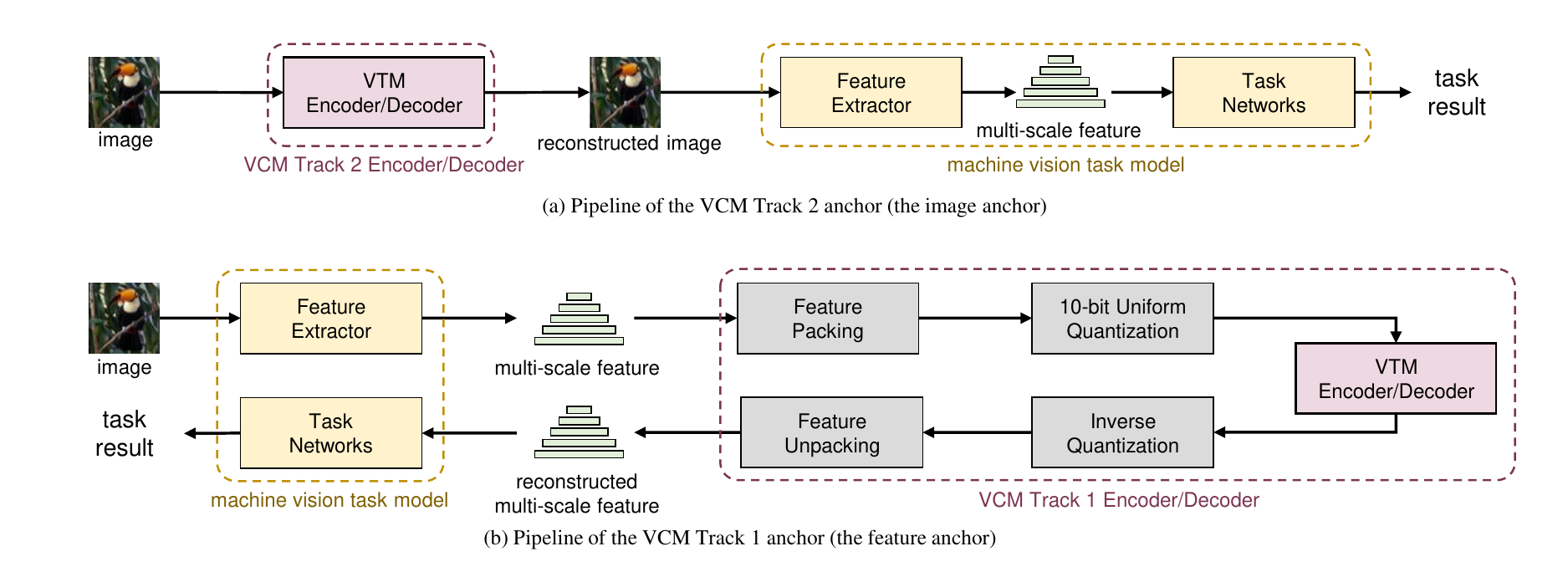}
\caption{Pipelines of the anchor models provided by MPEG-VCM corresponding to each of the tracks.}
\label{fig1}
\end{figure*}

\section{Introduction}\label{sec:introduction}
\IEEEPARstart{T}{raditional} image/video coding standards are optimized for the human visual system (HVS) under certain bitrate constraints. However, with the rise of machine learning applications, most images are now analyzed by machines rather than by humans. This led to the introduction of video coding for machines (VCM), which pursues high compression ratio while minimizing the performance degradation in machine analysis. After being initiated by MPEG (ISO/IEC JTC 1/SC 29) in 2019, VCM has growing attention and a number of related technologies have been explored so far. The MPEG-VCM activities have proceeded in two tracks: the feature compression track (Track1) and the image/video compression track (Track2). In Fig. \ref{fig1}, the pipelines of these tracks are shown with corresponding anchor models \cite{image-anchor, feature-anchor} provided by MPEG-VCM group. Call for proposal (CfP) for Track2 and Track1 have been issued in Apr. 2022 \cite{cfp2} and Apr. 2023, respectively.

Compared with image compression approaches like VCM Track2, feature compression (Track1) has two important merits \cite{req}:

\begin{itemize}
\item{With feature compression, privacy protection can be achieved as well as compression efficiency. Since the feature maps are extracted for machine vision, not HVS, it is difficult for humans to perceive objects from these feature maps.}
\item{In feature compression scenarios, feature extraction is performed on the encoder side so that the computational burden can be relieved on the decoder side. In other words, computational off-loading can be achieved.}
\end{itemize}

Zhang et al. \cite{zhang2021msfc} proposed a multi-scale feature compression method (MSFC), where a feature fusion module fuses the multi-scale feature maps into a single-scale feature map by aligning and concatenating them. On the decoder side, the multi-scale feature maps are reconstructed from the compressed single-scale feature map. In this way, the redundancy among the multi-scale feature maps can be effectively removed. Recently, Kim et al. \cite{kim2022} and Han et al. \cite{han2022} showed that introducing the versatile video coding (VVC) standard into MSFC could significantly improve compression performance. Despite these high compression performances, their methods are still sub-optimal because the conventional video coding standard was optimized only for natural images, not for the extracted features. Additionally, since the standard video codec is non-differentiable and thus is excluded in the training stage, their models cannot learn about coding noises at all. Zhang et al. \cite{zhang2022} proposed an end-to-end trainable model without using the standard video codec, resulting in some coding performance improvement. However, this method remains sub-optimal as it cannot eliminate redundancy among the multi-scale feature maps due to its layer-wise compression nature.

{As explained above, prior methods have some limitations such as incomplete end-to-end training, difficulty in eliminating redundancy between multi-scale feature maps, or using video codecs optimized only for natural images instead of feature maps. To overcome these limitations, we propose a new method called L-MSFC (Learnable Multi-Scale Feature Compression).} 

The main contributions of our paper are the following:

\begin{itemize}
\item We present a feature compression framework that can compactly integrate both multi-scale feature fusion and compression into an end-to-end trainable model.

\item We proposed an efficient way of combining the feature fusion and encoding processes. Our encoder interleaves the feature fusion and encoding process by transforming larger-scale feature into a latent representation and then fusing the result with the smaller-scale features iteratively.

\item We provide extensive experimental study and many informative analyses, including a trade-off between task-specific loss and reconstruction loss, the impact of distortions in each feature map layer on task performance, choice of feature mixing pathways in reconstructing multi-scale features at the decoder.

\item We suggest an additional evaluation metric for mission-critical VCM applications, where maintaining the highest task performance is the top priority.
\end{itemize}

\textcolor{red}{}

Our model was evaluated under the common test condition (CTC) of the feature compression track in MPEG-VCM \cite{ctc} and compared with the existing models \cite{kim2022, han2022, zhang2022} as well as the anchor models \cite{image-anchor, feature-anchor}. Experimental results show that our model outperforms all the existing models both in object detection and instance segmentation tasks, showing 98.22\% and 98.85\% BD-rate reductions against the feature anchor \cite{feature-anchor}. It is noteworthy that our model delivers near-lossless task performance with only 0.002-0.003\% of the uncompressed feature data size. Furthermore, our model exhibits a much faster encoding time compared with \cite{kim2022}. Note that the lightweight design of the encoder is highly desirable in many VCM applications where features are extracted and encoded at low-complexity edge devices.

\section{Related Works}
\subsection{MPEG-VCM Anchors}\label{subsec:MPEG-VCM anchor models}
Machine vision task models, such as Faster R-CNN \cite{ren2015faster} and Mask R-CNN \cite{he2017mask} for object detection and segmentation, typically include a feature extractor and task-specific networks. The feature extractor extracts meaningful features from the input image and the task networks then perform specific vision tasks based on the extracted features. Due to the separability of these architectures, both the input image and the extracted features can be subject to compression in the VCM context.

The standardization activities of MPEG-VCM are divided into two tracks, specifically the feature compression track (Track 1) and the image/video compression track (Track 2) \cite{cfe1}. Additionally, there is an anchor model provided for each track, namely feature anchor \cite{feature-anchor} and image anchor \cite{image-anchor}. These anchors serve as baselines of comparison against which related technologies can be easily evaluated. In the pipeline of the image anchor, as shown in Fig. \ref{fig1}(a), an input image is compressed by the VVC test model (VTM-12.0), and the machine vision task is performed with the compressed image. For the feature anchor, as shown in Fig. \ref{fig1}(b), features are first extracted from the input image and compressed through the VTM. Then, the decoded features are fed into task networks. Note that all the channels of the extracted features are spatially packed (i.e., tiled) into a single-channel picture for VTM encoding. 10-bit uniform quantization \cite{feature-anchor} is also performed before VTM encoding. The inverse process is applied to the output of the VTM decoder. As vision task models, Faster R-CNN X101-FPN and Mask R-CNN X101-FPN provided by Detectron2 \cite{wu2019detectron2} are used for object detection and instance segmentation, respectively. These models follow the original network architecture of \cite{ren2015faster} and \cite{he2017mask} except that ResNext \cite{xie2017aggregated}-based feature pyramid network (FPN) \cite{lin2017feature} is used as the feature extractor.

Fig. \ref{fig3} shows the architecture of FPN and the task networks of Faster/Mask R-CNN. It consists of a bottom-up and a top-down pathway. In the bottom-up pathway, initial multi-scale feature $C=\{c_2, c_3, c_4, c_5\}$ are extracted from an input image with the first four stages of the backbone, and they have strides of $\{4, 8, 16, 32\}$ pixels to the image. These feature maps are then passed to the top-down pathway with lateral connections maintaining the hierarchy. In the top-down pathway, each feature map of the hierarchical layers is combined with a larger one. Additionally, for detecting large objects, the smallest feature map $p_6$ is extracted from $p_5$ by subsampling. This final multi-scale feature $P=\{p_2, p_3, p_4, p_5, p_6\}$ is called the pyramidal feature or p-layer feature.

\begin{figure*}[!t]
\centering
\includegraphics[width=\linewidth]{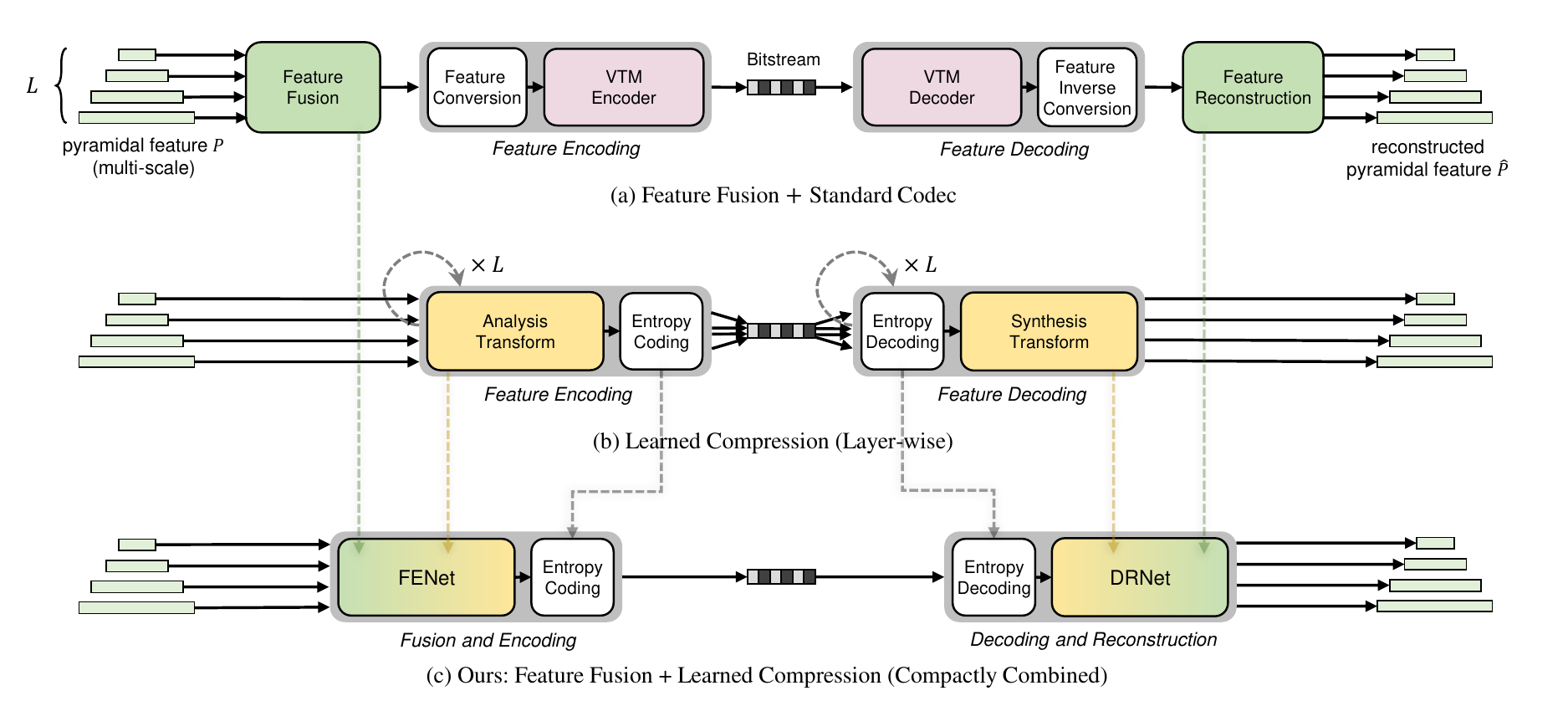}
\caption{Approaches for feature compression: (a) standard-codec-based multi-scale feature compression (S-MSFC), (b) learned layer-wise feature compression, (c) Ours: learned multi-scale feature compression (L-MSFC).}
\label{fig2}
\end{figure*}

The task networks comprise a set of neural networks which perform vision tasks based on the extracted pyramidal feature. As depicted in Fig. \ref{fig3}(b), region of interest (RoI) within an image that is likely to contain an object of interest is proposed by the region proposal network (RPN). Faster R-CNN applies RoI pooling to the found RoI, while Mask R-CNN utilizes RoI align. Finally, fully connected layers and task branches generate task results for the vision task. The dotted gray line represents the mask branch, which is unique to Mask R-CNN.

\begin{figure}[!t]
\centering
\includegraphics[width=\linewidth]{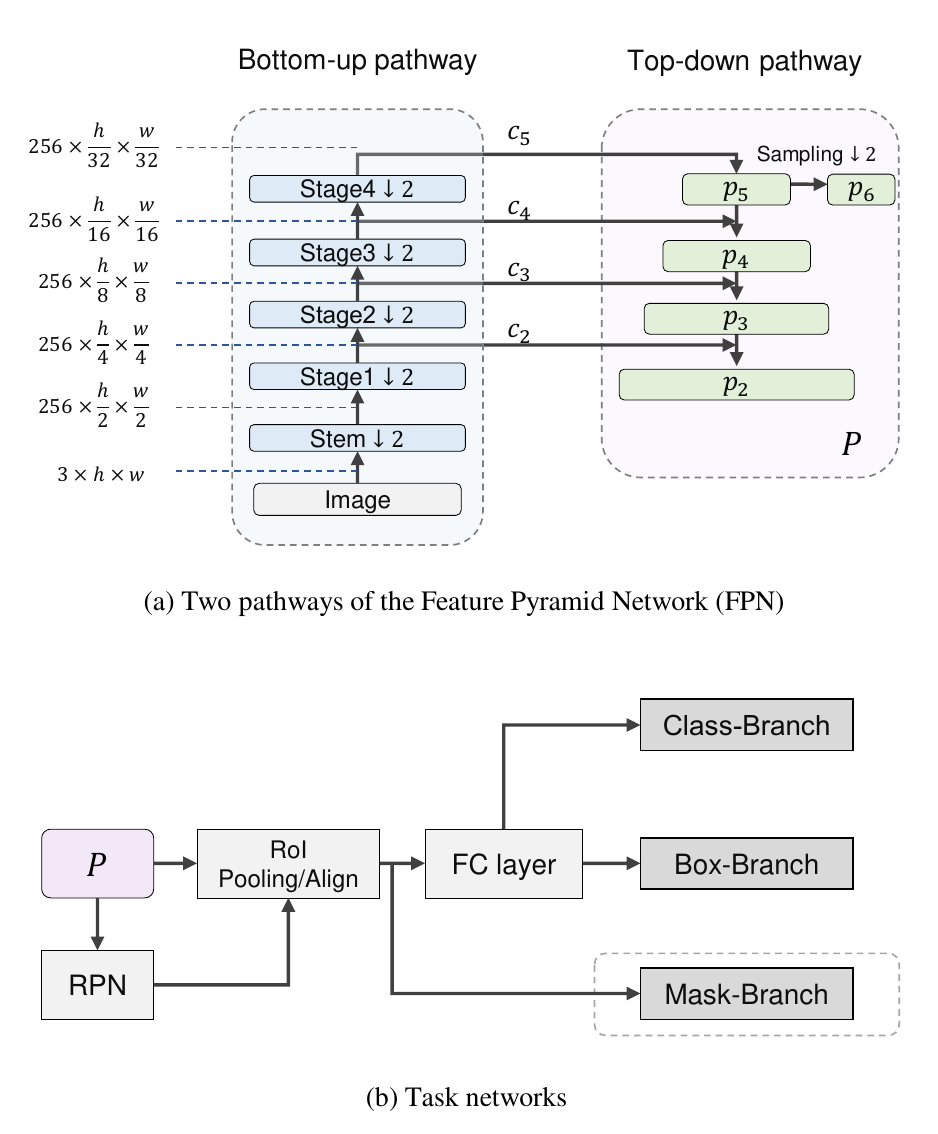}
\caption{The feature pyramid network (FPN) and task networks in the Mask R-CNN and the Faster R-CNN.}
\label{fig3}
\end{figure}

\begin{figure*}[!t]
\centering
\includegraphics[width=\linewidth]{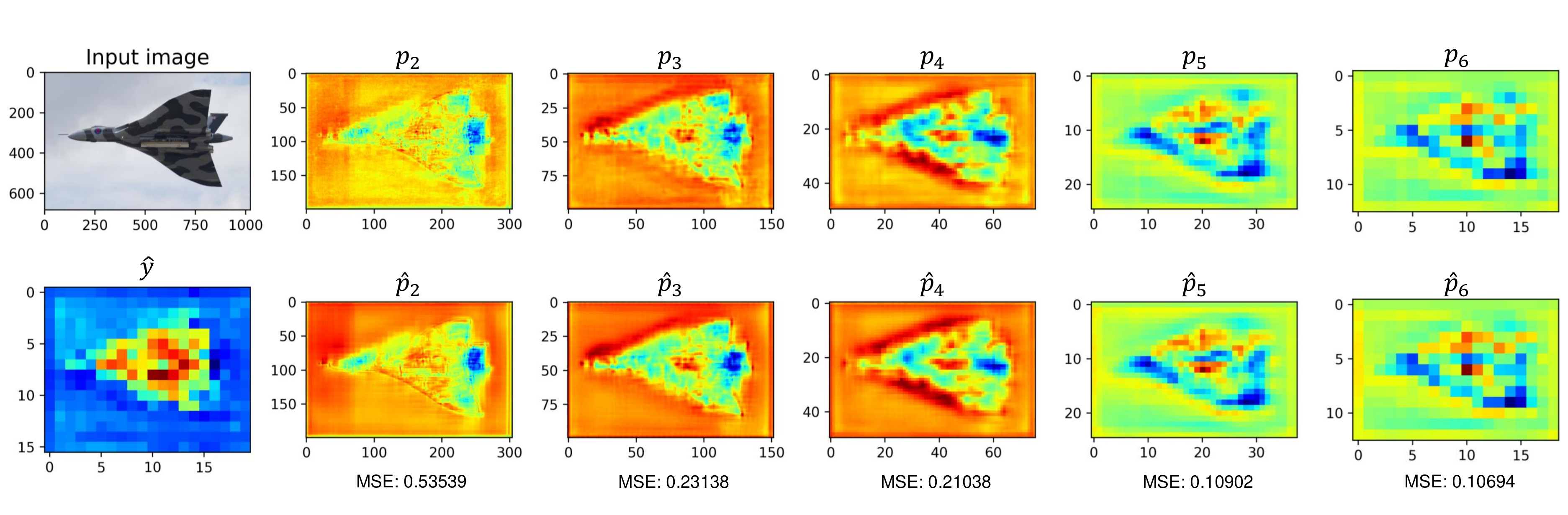}
\caption{{Visualization of the channel-averaged feature maps (by taking the average of values across all channels in each layer, followed by normalization into the [0, 1] range). The first row consists of an input image and the feature map of each layer extracted from the input image by the feature extractor of Faster R-CNN\cite{ren2015faster}. The second row displays fused latent representation $\hat{y}$ of the input feature maps and the reconstructed multi-scale feature maps with our proposed model. The mean squared error (MSE) for each reconstructed feature map is also shown.}}
\label{fig4}
\end{figure*}

\begin{figure}[!t]
\centering
\includegraphics[width=\linewidth]{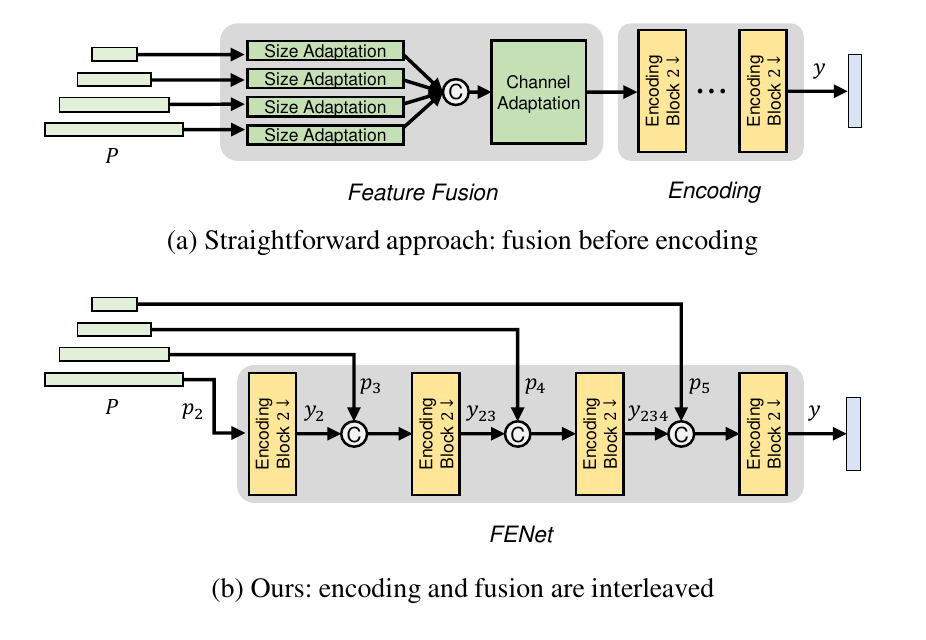}
\caption{Comparison between straightforward approach and ours for combining feature fusion and encoding.}
\label{fig5}
\end{figure}

\begin{figure*}[!t]
\centering
\includegraphics[width=\linewidth]{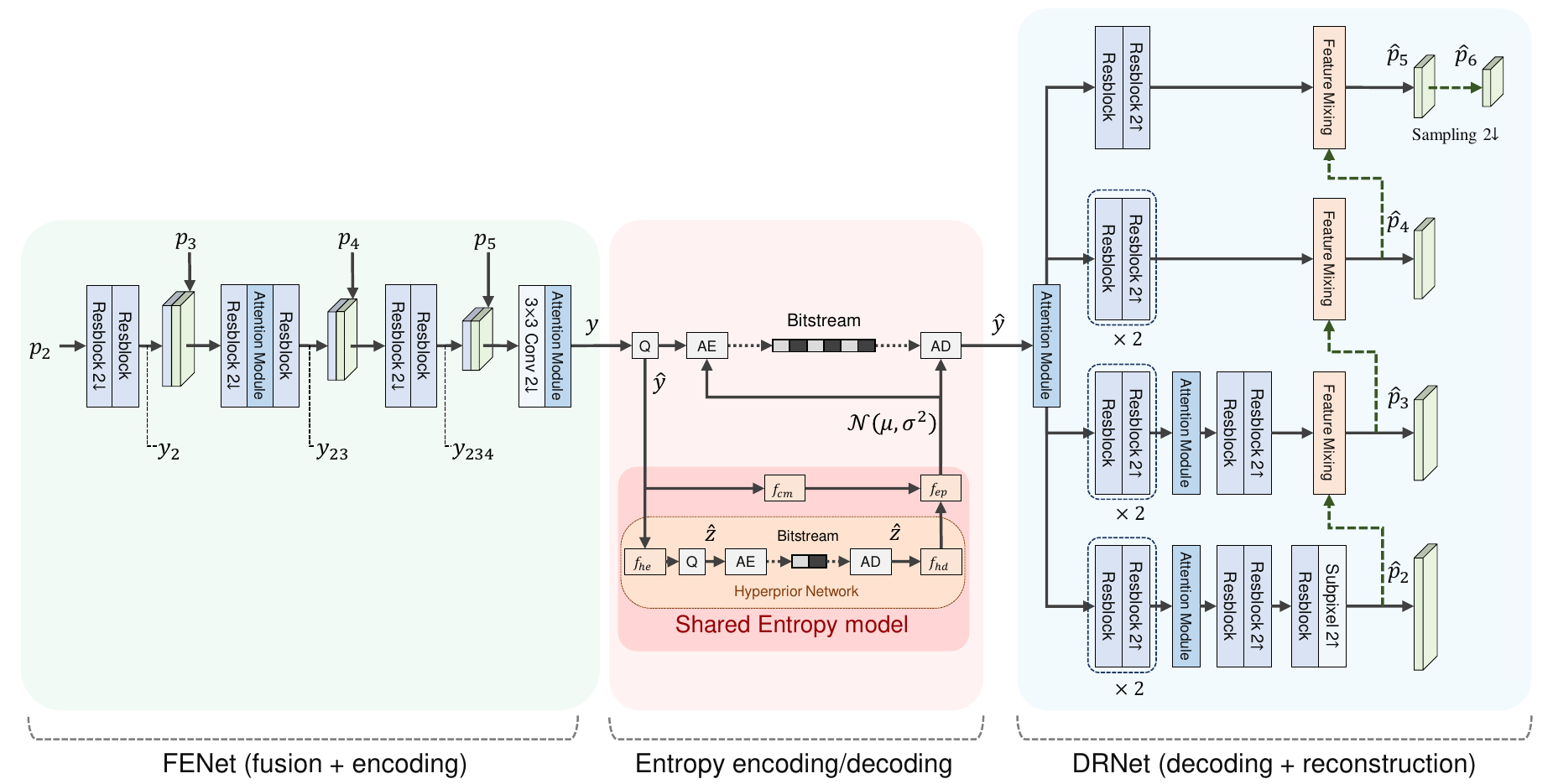}
\caption{Detailed architecture of the proposed model including FENet, DRNet, and the shared entropy model.}
\label{fig6}
\end{figure*}

\begin{figure*}[!t]
\centering
\includegraphics[width=\linewidth]{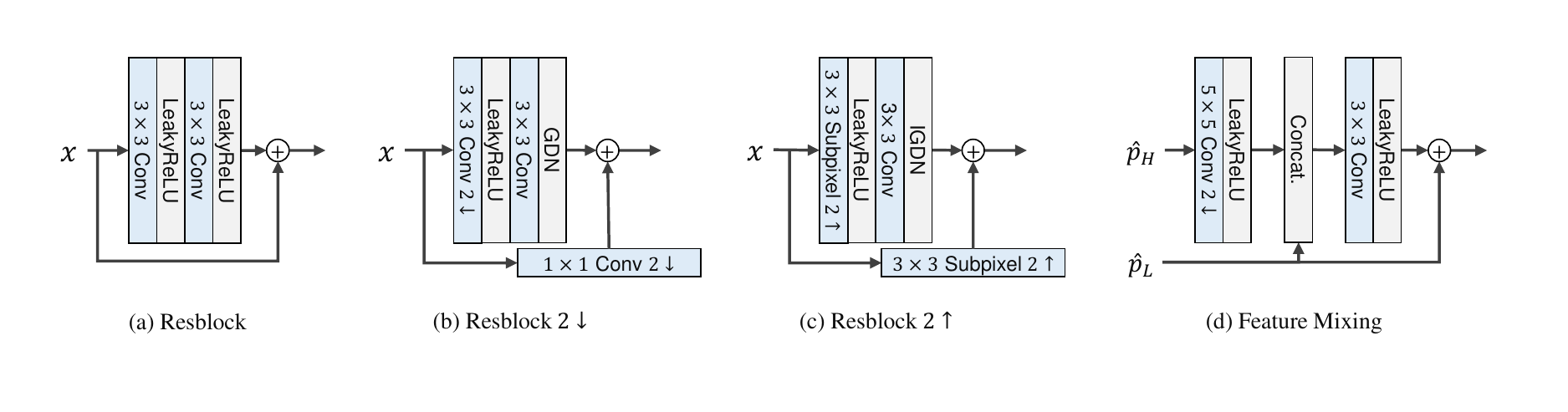}
\caption{Detailed architecture of sub-blocks in FENet and DRNet.}
\label{fig7}
\end{figure*}

\subsection{Feature Compression Models}
\label{subsec:vcmmodels}
Multi-scale feature compression method (MSFC) \cite{zhang2021msfc}, which was introduced in Section \ref{sec:introduction}, consists of a multi-scale feature fusion module (MSFF), single-stream feature codec (SSFC), and multi-scale feature reconstruction module (MSFR). The MSFF fuses the multi-scale feature maps extracted from the input image to generate a single-scale feature map. For this, the feature maps are aligned and concatenated with the same resolution as $p_5$. SE block \cite{hu2018squeeze} is then applied for re-weighting the channels of the fused feature map, based on the observation that each feature map has different importance for the vision task. The fused feature map is then compressed with n-bit uniform quantization. In the MSFR, the multi-scale feature maps are reconstructed through up-sampling lateral pathways and a bottom-up mixing pathway \cite{zhang2021msfc}.

Recently, \cite{kim2022} and \cite{han2022} showed that introducing the VTM into MSFC could  significantly improve compression performance, which we call standard codec-based MSFC (S-MSFC). More specifically, Han et al. \cite{han2022} replaced the sigmoid activation function in SSFC with PReLU \cite{he2015delving} and inserted VTM between the SSFC encoder and decoder. Kim et al. \cite{kim2022} replaced SSFC with the VTM and added a bottom-up pathway in MSFF module. These S-MSFC models achieved up to 96\% and 93\% BD-rate reductions against the VCM feature anchor\cite{feature-anchor} in object detection task. The overall architecture of S-MSFC models is shown in Fig. \ref{fig2}(a), where the feature conversion block includes the feature packing (which packs multi-channel features into a single frame) and 10-bit quantization, and the feature inverse conversion block includes feature unpacking and inverse quantization \cite{feature-anchor}.

Zhang et al. \cite{zhang2022} proposed an end-to-end trainable model, which adopts a learned image compression approach \cite{cheng2020}. They adjust the number of input channels of the image compression model and normalize the input feature maps between 0 and 1. Each of the multi-scale feature layers is then compressed by the model independently. Although this model does not contain the VTM and any feature fusion method, it achieved about 92\% BD-rate reductions against the feature anchor in instance segmentation task. This learned layer-wise compression approach is depicted in Fig. \ref{fig2}(b), where $L$ denotes the number of layers in the multi-scale feature.

Some researchers have sought other approaches for feature compression. For example, Lee et al. \cite{lee2022} applies principal component analysis (PCA) to the multi-scale feature maps. The major basis vectors of the feature maps are compressed with the VTM, and coefficients and mean coefficients are compressed with DeepCABAC \cite{wiedemann2020deepcabac}. Kang et al. \cite{kang2022, kang2023} exploits super-resolution (SR) techniques on the decoder side. For this, the feature maps are down-sampled to half the resolution and compressed with the VTM and then the low-resolution feature maps are upscaled with deep learning-based SR models to have the original resolutions. However, the feature fusion-based models \cite{kim2022, han2022} and the end-to-end model \cite{zhang2022} performed better than these models.

\subsection{Learned Image Compression}
\label{subsec:LIC}
Many works for learned image compression (LIC) have been proposed such as \cite{balle2018, minnen2018, lee2018, cheng2020, minnen2020, he2021checkerboard, lu2021transformer, zhu2021transformer, qian2021entroformer, zou2022devil, bai2022towards, hu2022coarse}. Ballé et al. \cite{balle2018} proposed autoencoder-based transforms and a hyperprior entropy model that estimates the probability distributions of the transformed latent representation with side information. This model became a key basis for subsequent studies, and deeper transforms and more accurate entropy models have been proposed rapidly. For example, Minnen et al. \cite{minnen2018} and Lee et al. \cite{lee2018} proposed autoregressive context models using PixelCNN \cite{van2016conditional} and fully connected layers, respectively. These context models use already decoded elements of the latent representation as additional context information for the entropy model. These models outperformed the HEVC \cite{sullivan2012overview} intra-coding. Cheng et al. \cite{cheng2020} improved the transform networks with residual blocks \cite{he2016deep} and simplified attention modules. They also used the Gaussian mixture model (GMM) for estimating more accurate probability distributions of the latent representation. This model showed comparable performance with VVC \cite{bross2021overview} intra-coding. More advanced transforms and entropy models have been proposed recently \cite{minnen2020, he2021checkerboard, lu2021transformer, zhu2021transformer, bai2022towards, zou2022devil, qian2021entroformer, hu2022coarse}, some of which use vision transformer \cite{dosovitskiy2021an}.

\section{Proposed Methods}\label{sec:proposedmethods}
\subsection{Overview}\label{subsec:overview}
We provide a novel architecture that efficiently integrates multi-scale feature fusion with a learnable compressor. Fig. \ref{fig2} illustrates the structural differences of our model from the previous methods. Our model differs from S-MSFC approaches \cite{kim2022, han2022} in a sense that it does not include any conventional video codec component and hence allows end-to-end training. Unlike the learned layer-wise compression approach \cite{zhang2022}, our model exploits the feature fusion technique. As shown in the first row of Fig. \ref{fig4}, the multi-scale features exhibit high spatial redundancy across layers. Hence, it is highly desirable to introduce some effective way to remove such redundancy such as feature fusion (e.g., \cite{zhang2021msfc}) or feature resizing (e.g., \cite{kang2022, kang2023}).

The detailed architecture of the proposed model is shown in Fig. \ref{fig6}, where FENet {(Fusion and Encoding Network)} integrates the feature fusion with compression processes and DRNet {(Decoding and Reconstruction Network)} combines the decompression and multi-scale feature reconstruction processes. We will explain the details of the components and training objectives in the following subsections.

\subsection{Feature Fusion and Encoding Network (FENet)}\label{subsec:ffen}

One possible approach to integrate a feature fusion network with a trainable compressor is to simply concatenate them in a sequential manner (as shown in Fig. \ref{fig5}(a)). Though it is straightforward, this combination would result in many redundant computations; for example, the size adaptation is performed both in the fusion network and the encoding network. In this paper, we propose a more compact model, i.e., FENet, that can perform the two functions with a much smaller number of computational components. Fig. \ref{fig5}(b) shows the proposed FENet architecture, where the encoded latent of the lower layer feature maps is successively fused with the higher layer feature maps. More specifically, the lowest-layer feature map $p_{2}$ is first encoded into a latent representation $y_{2}$ through the first encoding block. The latent $y_{2}$ is then fused with the higher-layer feature map $p_{3}$ before fed into the second encoding block. Since the spatial dimension of $y_{2}$ and $p_{3}$ are all the same, we can simply use channel-wise concatenation to fuse the latent and the feature map. The second encoding block then encodes the concatenated input \{$y_{2}$, $p_{3}$\} to obtain $y_{23}$. After repeating this process successively, we can get the final encoded latent $y=y_{2345}$, which contains the accumulated information from $p_{2}$ to $p_{5}$. Note that FENet in Fig. \ref{fig5}(b) would have similar complexity to the encoding part of the straightforward model in Fig. \ref{fig5}(a).

\begin{table*}[htb!]
\centering
    \caption{Detailed information of the modules in the entropy model}
    \begin{tabular}{c c c c}
    \toprule
    \addlinespace[1ex]
    \textbf{Hyper Encoder ($f_{he}$)} & \textbf{Hyper Decoder ($f_{hd}$)} & \textbf{Context Model ($f_{cm}$)} & \textbf{Entropy Parameters ($f_{ep}$)} \\
    \midrule
        $3 \times 3$ Conv         N                &
        $3 \times 3$ Conv         N                &
        5$\times$5 Masked Conv  N                &
        1$\times$1 Conv         10N/3            \\
    
        Leaky ReLU                              & 
        Leaky ReLU                              &    
                                                & 
        Leaky ReLU                              \\
    \addlinespace[1ex]
        $3 \times 3$ Conv         N               & 
        $3 \times 3$ Subpixel     N     2$\uparrow$ &
                                                & 
        1$\times$1 Conv         8N/3            \\
    
        Leaky ReLU                              & 
        Leaky ReLU                              & 
                                                & 
        Leaky ReLU                              \\
    \addlinespace[1ex]
        $3 \times 3$ Conv         N   2$\downarrow$ & 
        $3 \times 3$ Conv         3N/2            & 
                                                & 
        1$\times$1 Conv         2N              \\
    
        Leaky ReLU                              & 
        Leaky ReLU                              & 
                                                & 
                                                \\
    \addlinespace[1ex]
        $3 \times 3$ Conv        N                & 
        $3 \times 3$ Subpixel    3N/2   2$\uparrow$ &
                                                & 
                                                \\
        Leaky ReLU                              & 
        Leaky ReLU                              & 
                                                & 
                                                \\
    \addlinespace[1ex]
        $3 \times 3$ Conv        N    2$\downarrow$ & 
        $3 \times 3$ Conv        2N               &
                                                & 
                                                \\
    \bottomrule
    \end{tabular}
    \label{table:entropy model details}
\end{table*}

As shown in Fig. \ref{fig6}, the first three encoding blocks in FENet are implemented with the residual blocks \cite{he2016deep}, and in the second block, the simplified attention module is also used. For the last encoding block, a convolution layer and the simplified attention module \cite{cheng2020} are used. The output of FENet, $y$, is then entropy-coded by an  arithmetic encoder (AE) based on the shared entropy model (Section \ref{subsec:entropy model}). The detailed architectures of the sub-blocks of {FENet} are shown in Fig. \ref{fig7}(a)-(b). The residual block ("Resblock") includes two convolutional layers that have $3 \times 3$ kernels with stride 1, LeakyReLU activations \cite{xu2015empirical}, and the residual connection \cite{he2016deep}. For the down-sampling residual block ("Resblock $2\downarrow$"), the stride of the first convolutional layer is set to 2, and the last LeakyReLU activations are replaced with the generalized divisive normalization (GDN) \cite{balle2018}. In addition, a $1\times1$ convolutional layer with stride 2 is applied to the residual connection rather than the identity function.

Note that GDN was first proposed in \cite{balle2015density} as a parametric nonlinear transformation that is well-suited for Gaussianizing data from natural images. It has been not only commonly used in learned image compression but also adopted in a learned feature compression model \cite{zhang2022}. Since the input feature maps of the FENet are also extracted features from a natural image, GDN could also be applied to our proposed encoder as was in LIC encoders.




\subsection{Feature Decoding and Reconstruction Network (DRNet)}\label{subsec:fdrn}
Referring to Fig. \ref{fig6}, the encoded bitstream of the latent representation is entropy-decoded by {an arithmetic decoder (AD) based on the shared entropy model.} In {DRNet}, the simplified attention module \cite{cheng2020} is applied to the decoded latent representation $\hat{y}$ and then the output is passed to the following layer-wise branches to reconstruct the multi-scale feature maps, where the reconstructed feature maps are denoted by $\hat{p}_2, \hat{p}_3, \hat{p}_4$ and $\hat{p}_5$ in the figure.

Each branch includes several residual blocks where the number of blocks depends on the resolution of the corresponding feature map. {We use the deeper network for the higher-resolution feature map since they require more upsampling blocks.} For $\hat{p_2}$ and $\hat{p_3}$, the simplified attention module is inserted between the intermediate residual blocks. The reconstructed lower-layer feature map is fed into the feature mixing block of the upper-layer branch to help reconstruct the upper-layer feature map.

The detailed architectures of the up-sampling residual block ("Resblock $2\uparrow$") and the feature mixing block ("Feature Mixing") are shown in Fig. \ref{fig7}(c)-(d). The up-sampling residual block replaces the first layer of the residual block (Fig. \ref{fig7}(a)) with a $3 \times 3$ subpixel convolution \cite{shi2016real} with stride 2, and the last activation with the inverse GDN (IGDN) \cite{balle2018}. For the residual connection, $3 \times 3$ subpixel convolution with stride 2 is also applied. The feature mixing block takes a higher resolution feature map $\hat{p}_{H}$ and a lower resolution feature map $\hat{p}_{L}$. The $\hat{p}_{H}$ goes through a $5\times5$ convolutional layer with stride 2 and then the output is concatenated with the $\hat{p}_{L}$. The concatenated feature map is mixed in the following $3 \times 3$ convolutional layer with stride 1. The output is added to the original $\hat{p}_{L}$ for residual connection.

\subsection{Entropy Model}\label{subsec:entropy model}
{To perform entropy coding effectively, precise estimation of the probability distribution of symbols to be coded is required. In the context of learned image compression, the assumed probability model is often called an entropy model, and the model parameters are estimated through neural networks. Several entropy models have been proposed such as \cite{balle2017, balle2018, minnen2018, minnen2020}. For our shared entropy model, we adopted \cite{minnen2018}, which exploits both an autoregressive context model and hyperprior networks. The operation diagram of the shared entropy model is shown in Fig. \ref{fig6} where $f_{cm}$ refers to the autoregressive context model which extracts context information from previously decoded elements of the latent representation, and $f_{hd}$ refer to the hyperprior encoder and decoder, respectively. $f_{ep}$ denotes the entropy parameter module that estimates the entropy model parameters from the outputs of the context model $f_{cm}$ and the hyperprior decoder $f_{hd}$. In addition, "Q" is uniform quantization, and "AE" and "AD" denote arithmetic encoder and decoder, respectively. Note that the quantization operations in the figure are replaced with adding uniform noise in the training stage as in \cite{balle2018}.}

\begin{figure}[!t]
\centering
\includegraphics[width=\linewidth]{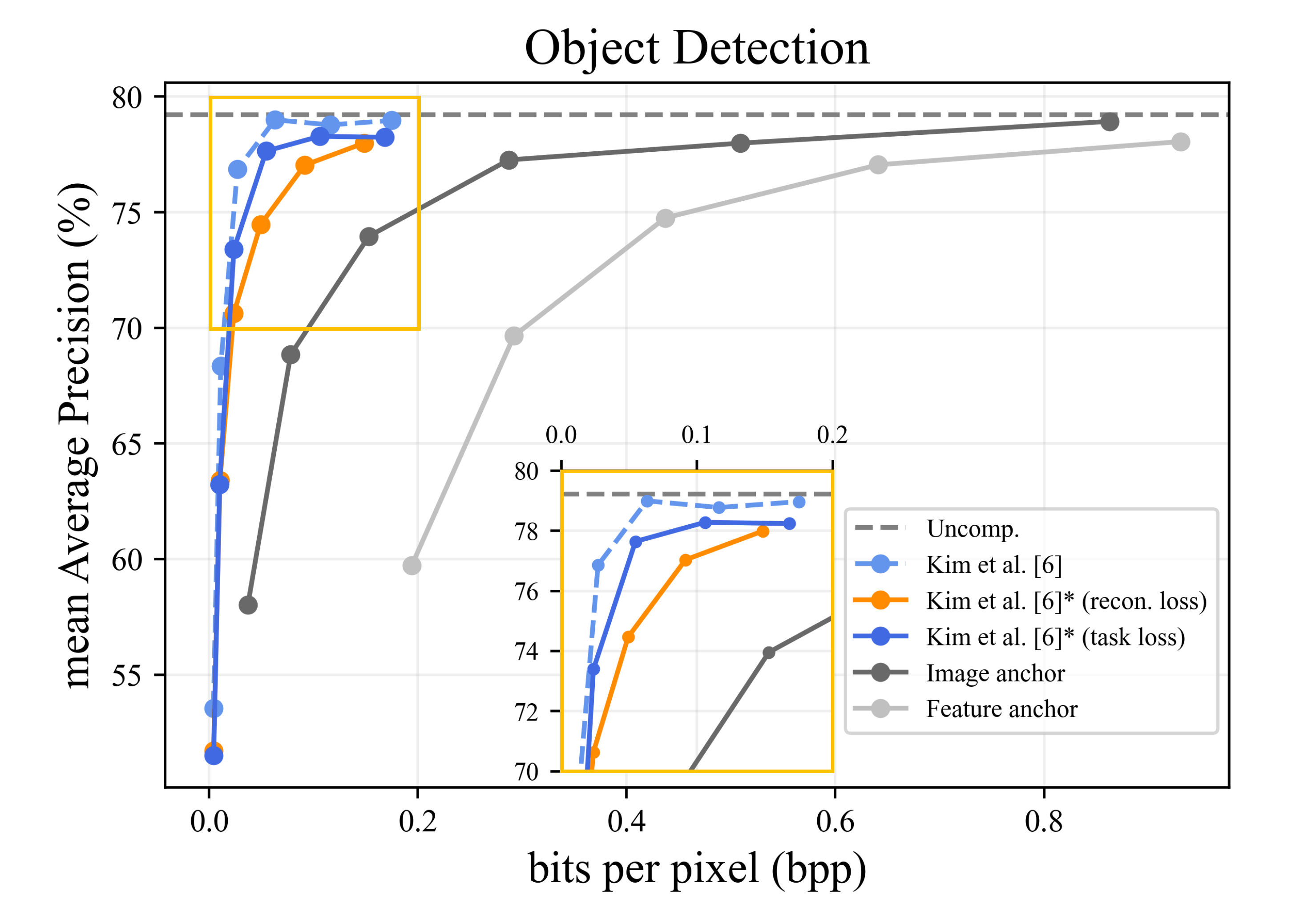}
\caption{Performance comparison between task-loss-trained model \cite{kim2022} and reconstruction-loss-trained one. The models reproduced by us are denoted with "*". The performance reported in the original reference \cite{kim2022} is also shown with a dashed curve.}
\label{fig8}
\end{figure}

\begin{figure*}[!t]
\centering
\includegraphics[width=\linewidth]{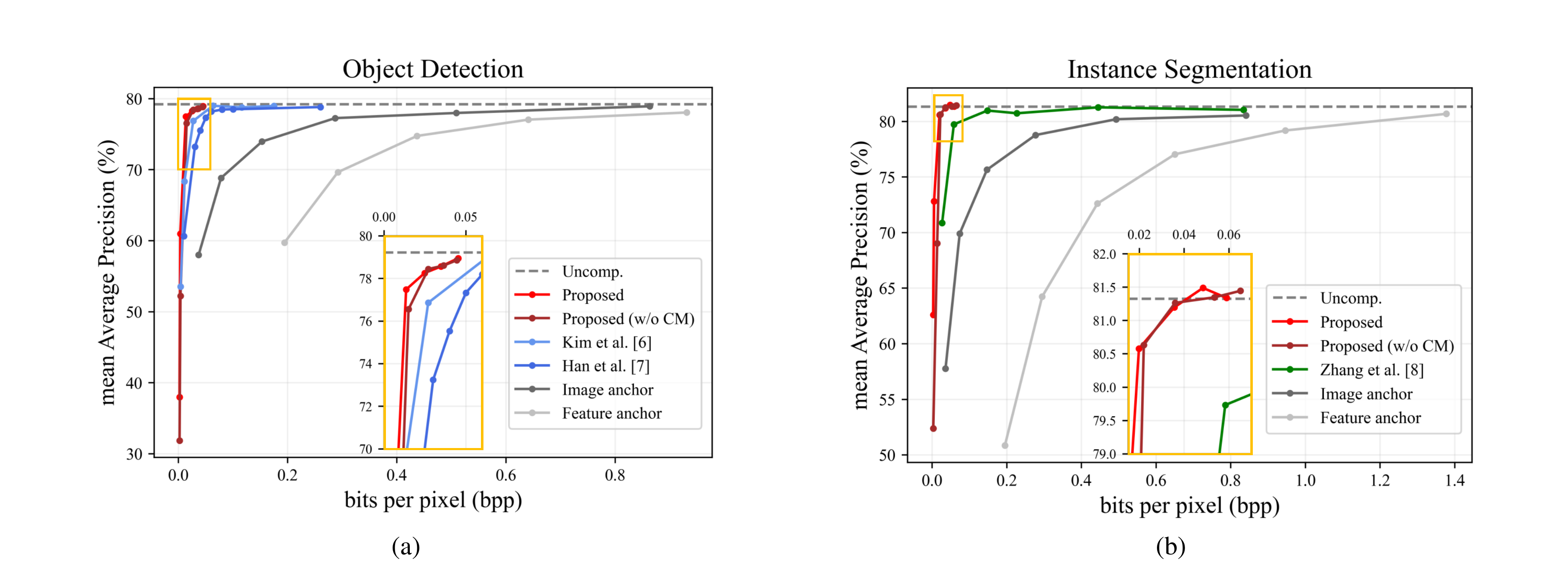}
\caption{Rate-performance curves of the proposed models and previous methods. {(a): for object detection task, and (b): for instance segmentation task.}}
\label{fig9}
\end{figure*}

We use CompressAI's implementations \cite{begaint2020compressai} of the joint autoregressive entropy model \cite{minnen2018}. Detailed information on the modules in the entropy model is shown in Table \ref{table:entropy model details}. Each entry of this table specifies the kernel size of the layer, which is represented as $K\times~K$, the layer type, which is a convolutional layer ("Conv"), or masked convolution ("Masked Conv") \cite{van2016conditional}, or subpixel convolution ("Subpixel") \cite{shi2016real}, the number of output channels, and finally the activation layer. In this table, "$\uparrow$" and "$\downarrow$" mean up-sampling and down-sampling, respectively.

\subsection{Training Loss}\label{subsec:otherdetails}
\begin{table*}[th]
    \small
    \centering
     \caption{{BD-rate comparisons}}
    \resizebox{\textwidth}{!}{
    \begin{tabular}{c c c c c c c c c c}
    \toprule
    \multirow{2}{*}{\textbf{Model}}     & 
    \multirow{2}{*}{\textbf{Task loss}} &
    \multirow{2}{*}{\textbf{VTM}}       &&&
    \multirow{1}{*}{\textbf{Object Detection}} &&&  
    \multirow{1}{*}{\textbf{Instance Segmentation}}&     \\
    \cmidrule(lr){5-7} 
    \cmidrule(lr){8-10}
     &  & & vs. &
    \textbf{Image Anchor}   &
    \textbf{Feature Anchor} &
    \textbf{Kim et al. \cite{kim2022}} & 
    \textbf{Image Anchor}   &
    \textbf{Feature Anchor} &
    \textbf{Zhang et al. \cite{zhang2022}} \\
    \midrule
    
    Image Anchor \cite{image-anchor}&   N  &    Y   &&
    0.00\%      &
    -69.86\%    &
    650.57\% 
    &  
    0.00\%      &
    -79.52\%    &
    212.59\%    \\
        
    Zhang et al. \cite{zhang2022}   &   N   & N &&
    -           &
    -           &
    -           &
    -68.01\%    &  
    -92.30\%    &
    0.00\%           \\
    
    Han et al. \cite{han2022}   &   Y   &   Y   &&
    -77.80\%    &
    -93.19\%    & 
    68.37\%     &
    -           &
    -           &
    -           \\

    Kim et al. \cite{kim2022}   &   Y   &   Y   &&
    -86.68\%    &
    -96.01\%    &
    0.00\%      & 
    -           &
    -           &
    -           \\
    \midrule
    Proposed (w/o CM)    &   N   &   N   &&
    -91.06\%    &
    -97.38\%    & 
    -28.96\%    &
    -85.10\%    &
    -97.21\%    &
    -60.47\%    \\
    
    Proposed    &   N  &    N   &&
    \textbf{-93.95}\%&
    \textbf{-98.22}\%&
    \textbf{-52.23}\%&
    \textbf{-95.79}\%&
    \textbf{-98.85}\%&
    \textbf{-84.58}\%\\
    \bottomrule
\end{tabular}
}
     \label{table:BD-rate gains}
\end{table*}

{Feature compression models can be trained with two different loss types. \cite{kim2022} and \cite{han2022} used a task-specific loss to maximize the coding performance while \cite{zhang2022} used a reconstruction loss which could improve generalization over different tasks. Fig. \ref{fig8} shows the experimental comparison between the task loss and the reconstruction loss. The model we used for this experiment is our reproduced version of \cite{kim2022}. As seen in the figure, both losses achieve superior performance compared with the MPEG-VCM anchor models \cite{feature-anchor, image-anchor}. Specifically, BD-rate gains against the image anchor are 76.12\% with the task loss and 81.51\% with the reconstruction loss. As expected, the task-specific loss showed a higher gain compared with the reconstruction loss. The training time for the task loss, however, was roughly four times longer than that for the reconstruction loss. If task-specific optimization is allowed and a higher compression efficiency is priority, a task loss could be a better choice. On the other hand, if a task loss is inaccessible or task-independent optimization is desirable for better generalization, reconstruction loss could be used instead. Since the main focus of this paper is to propose a model that can achieve sufficiently high performance even without a task-specific loss, we use reconstruction loss for training our model.}

{Specifically}, we used the rate-distortion loss $L$ defined by:
\begin{equation}
\label{eq:2}
    L= R+ \lambda D_{total}
\end{equation} where $R$ denotes the rate loss, $D_{total}$ denotes the distortion between the original and the reconstructed feature maps, and $\lambda$ is a Lagrangian multiplier that determines the trade-off between rate and distortions. As the value of $\lambda$ increases, the quality of reconstruction also increases. We train multiple models with different $\lambda$'s to test different quality levels. As was done in \cite{minnen2018}, we estimate the rate $R$ by (\ref{eq:3}), where $p_{\hat{y}|\hat{z}}(\hat{y}|\hat{z})$ and $p_{\hat{z}}(\hat{z})$ denote the probability distributions for the latent representation $\hat{y}$ and the hyperprior $\hat{z}$, respectively.
\begin{equation}
\label{eq:3}
  R = \mathbb{E}_{{\hat{y} \sim q_{\hat{y}}, \hat{z}} \sim q_{\hat{z}}}[-log_{2}{p_{\hat{y}|\hat{z}}(\hat{y}|\hat{z})} -log_{2}{p_{\hat{z}}(\hat{z})}]
\end{equation} We define the total distortion $D_{total}$ as a weighted combination of layer-wise distortion as shown in (\ref{eq:4}), where $D(p_i, \hat{p}_i)$ denotes mean squared error (MSE) between the original and the reconstructed feature map of $i^{th}$ layer and $w_{i}$ is the relative weight on the $i^{th}$ layer distortion $D(p_i, \hat{p}_i)$.
\begin{equation}
\label{eq:4}
  \begin{aligned}
    D_{total} = \sum_{i=2}^{6}{w_{i}D(p_i, \hat{p}_i)}
    \quad with \quad \sum_{i=2}^{6}{w_{i}=1}
  \end{aligned}
\end{equation}

\section{Experiments}\label{sec:experiments}
\subsection{Training and Implementation Details}\label{subsec:training and implementation details}
\subsubsection{Datasets}
For training, we randomly selected about 90,000 images with a resolution higher than 512$\times$512 from the OpenImages (V6) \cite{OpenImages} training set. Then, we randomly cropped the patches with a 512$\times$512 size from that images. 
\subsubsection{{Models}}

{The proposed model used the same entropy model as in \cite{zhang2022}, which includes the autoregressive context model \cite{minnen2018} that uses the already decoded latent elements for encoding subsequent ones to achieve higher coding efficiency. However, the context model (CM) involves sequential dependency resulting in a significant increase in decoding complexity. Taking this trade-off into consideration, we evaluate our model in two settings: with and without the context model (w/o CM).} The numbers of output channels of the layers in {FENet and DRNet} are set to a fixed number $N$ except for the last layer of the branches in {DRNet}. The specific value of $N$ is determined according to $\lambda$ values and whether the context model is used or not, as summarized in Table \ref{table:n}.

\subsubsection{{Training settings}}
{To obtain results for various rate points, we trained each of the models using 6 different lambda values where $\lambda_{1}$ to $\lambda_{6}$ set to 0.0125, 0.025, 0.125, 0.25, 0.375 and 0.5. The weight on the distortion for the $i^{th}$ layer, $w_{i}$, is set to 0.2 for all layers.} We used Adam optimizer with default settings of Pytorch \cite{paszke2019pytorch} with a batch size of 4. {The learning rate is initially set to 1e-4 and is then halved whenever the validation loss reaches a plateau until it is higher than 5e-6.} Finally, the models are fine-tuned with the original sizes of images with a batch size of 1 and a learning rate of 1e-5 for 200,000 iterations.

\begin{table}[t!]
    \caption{The number of channels $N$ for {FENet and DRNet in the proposed models.}}
    \small
    \centering
    \resizebox{0.8\linewidth}{!}{
    \begin{tabular}{c c c}
    \toprule
    $\lambda$   & $\lambda_{1}$, $\lambda_{2}$, $\lambda_{3}$ & $\lambda_{4}$, $\lambda_{5}$,$\lambda_{6}$ \\

    \midrule
    Proposed          &  N=128  &  N=192    \\
    Proposed (w/o CM) &  N=192  &   N=192   \\
    \bottomrule
     \end{tabular}
     }
     \label{table:n}
\end{table}

\begin{table}[ht]
    \small
    \centering
     \caption{The values of bpp, $D_{total}$, and mAP obtained with the proposed model. There is also a case for the uncompressed feature}
    \resizebox{\linewidth}{!}{
    \begin{tabular}{l ccc ccc}
    \toprule
                                                            & 
    \multicolumn{3}{c}{Object Detection}               & 
    \multicolumn{3}{c}{Instance Segmentation}                    \\
    \cmidrule(lr){2-4} \cmidrule(lr){5-7}
    $\lambda$~(quality level) & bpp & $D_{total}$ & mAP& bpp& $D_{total}$ &  mAP  \\ 
    \midrule
        $\lambda_{1}=0.0125$      &   0.0019  & 0.654    &   37.977  &   0.0029  &   0.932    &   65.594     \\
        $\lambda_{2}=0.025$       &   0.0033  & 0.524    &   60.999  &   0.0053  &   0.748    &   72.819     \\
        $\lambda_{3}=0.125$       &   0.0135  & 0.306    &   77.484  &   0.0198  &   0.470    &   80.578     \\
        $\lambda_{4}=0.25$        &   0.0248  & 0.236    &   78.256  &   0.0356  &   0.422    &   81.198    \\ 
        $\lambda_{5}=0.375$       &   0.0348  & 0.194    &   78.562  &   0.0484  &   0.278    &   81.488     \\
        $\lambda_{6}=0.5$         &   0.0453  &   0.176    &   78.953    &   0.0589  &   0.260    &   81.339    \\
        \midrule
        Uncompressed&   841.940 & -   &   79.225    &   830.686 &  -   &81.326   \\
        \bottomrule
        \end{tabular}
        }
     \label{table:rate-points}
\end{table}

\begin{table}[t!]
    \small
    \centering
     \caption{${R_{NL}}$ and ${CR_{NL}}$ Comparison}
    \resizebox{\linewidth}{!}{
    \begin{tabular}{l ccc ccc}
    \toprule
                                                            & 
    \multicolumn{2}{c}{Object Detection}               & 
    \multicolumn{2}{c}{Instance Segmentation}                    \\
    \cmidrule(lr){2-3} \cmidrule(lr){4-5}
    Model & $\boldsymbol{R_{NL}\downarrow}$ & $\boldsymbol{CR_{NL}\uparrow}$ & $\boldsymbol{R_{NL}\downarrow}$ & $\boldsymbol{CR_{NL}\uparrow}$  \\ 
    \midrule
        Feature Anchor \cite{feature-anchor}    &   1.128  &   746 &   1.326  &   635     \\
        Image Anchor \cite{image-anchor}      &     0.735  &   1,146 &   0.712  &   1,183      \\
        Zhang et al. \cite{zhang2022}        &       -      &     -     &   0.070  &   12,004      \\
        Han et al. \cite{han2022}            &   0.070     & 12,079   &     -    &       -         \\ 
        Kim et al. \cite{kim2022}            &   0.040     & 21,187  &      -    &       -         \\
        \midrule
        Proposed (w/o CM)                    &  \textbf{0.027}      & \textbf{31,505}   &   0.022  &   38,680          \\
        Proposed                             &   0.031     &   27,555 &   \textbf{0.019}   &   \textbf{44,426}      \\
        \bottomrule
        \end{tabular}
        }
     \label{table:crnl-points}
\end{table}

\subsection{Experimental Results}\label{subsec:experimental results}
\subsubsection{Evaluation Conditions}
For fair evaluations of the VCM proposals, MPEG-VCM provides common test conditions (CTC) [9]. The CTC describes evaluation metrics, test datasets, target vision tasks, models for the vision tasks, and other details. For all the experiments in this paper, we follow the conditions described in the CTC. As defined in the CTC, the overall coding performance is measured with BD-rate, where bits per pixel (bpp) is used as rate measure and mean average precision (mAP) with intersection over union (IoU) threshold 0.5 is used as task performance. Note that the CTC calculates the bpp as the bitstream size divided by the number of pixels in the input image.

We also propose an additional metric for mission-critical VCM applications, where maintaining the highest task performance is the top priority. In compression applications for HVS, image quality degradation does not necessarily undermine the service itself. In VCM applications, however, degradation in task performance is unacceptable if it could result in some emergency cases such as car accidents and so on. In this context, we propose near-lossless bitrate $R_{NL}$, which is defined as the bitrate required to achieve less than 1\% task performance degradation of the original performance without compression. Furthermore, to measure the compression ratio at the near-lossless bitrate, we define $CR_{NL}$ as follows:
\begin{equation}
\label{eq:5}
  \begin{aligned}
    CR_{NL} = R_{0}/R_{NL},
  \end{aligned}
\end{equation}
where $R_{0}$ is the bitrate of uncompressed feature maps. Note that $CR_{NL}$ indicates how many times less bitrate is needed compared to the original data to achieve near-lossless task performance. We evaluate our model and reference models with not only BD-rate but also the proposed metrics. We believe $R_{NL}$ or $CR_{NL}$ will be helpful in evaluating VCM technologies.

\begin{figure*}[!t]
\centering
\includegraphics[width=\linewidth]{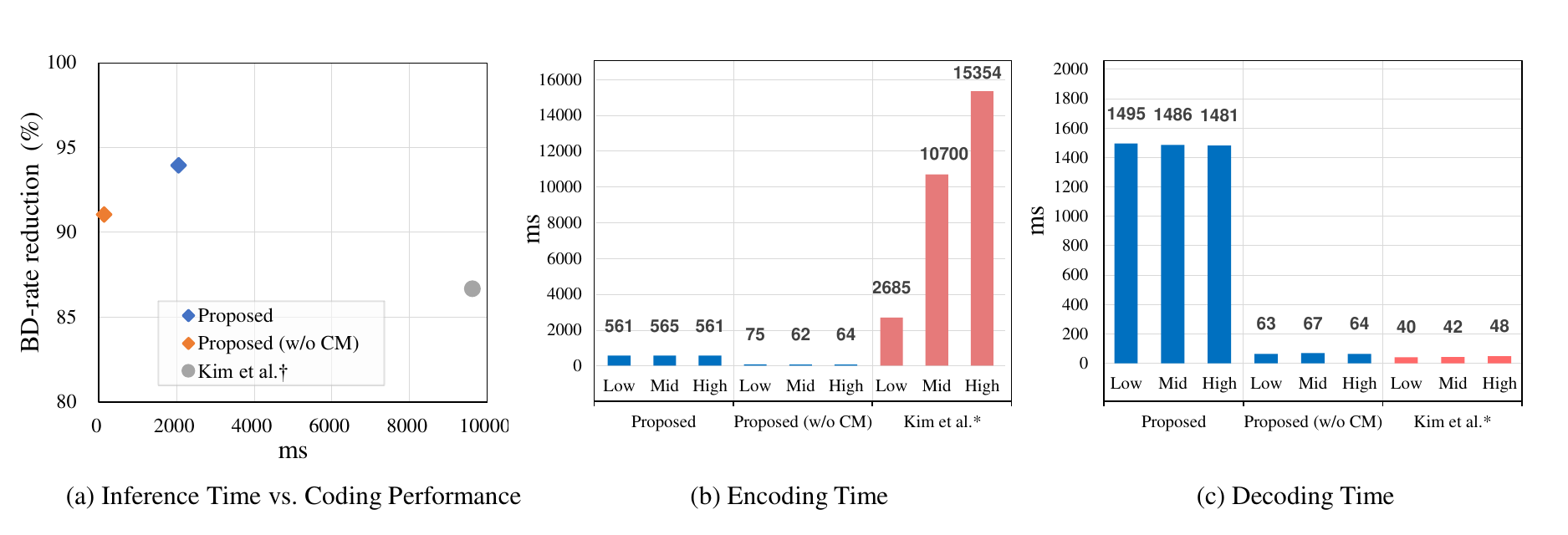}
\caption{Comparison of proposed models with \cite{kim2022} in terms of complexity. (a) shows average inference times and bit-rate reduction compared to VCM image anchor. (b) and (c) show encoding and decoding times of proposed models and \cite{kim2022} at various quality levels, where the "High" level model has the most bit allocation and highest task performance. "$\dag$" in (a) indicates that the compression performance is obtained from the original model \cite{kim2022}, while the inference time is measured in the reproduced model.}
\label{fig10}
\end{figure*}

\begin{figure}[!t]
\centering
\includegraphics[width=\linewidth]{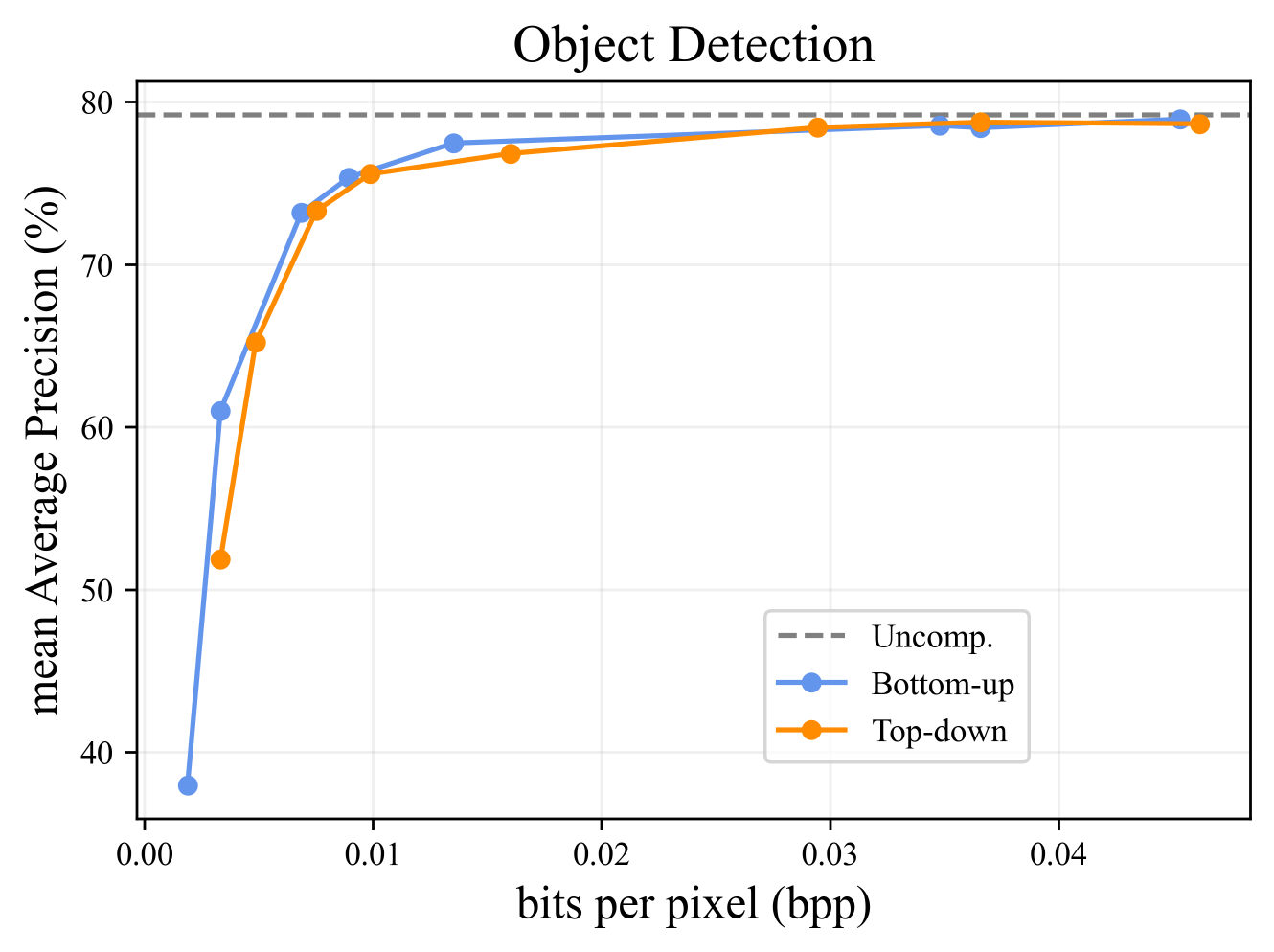}
\caption{Performance comparison between using top-down and bottom-up pathways for feature mixing.}
\label{fig11}
\end{figure}

\begin{figure}[!t]
\centering
\includegraphics[width=\linewidth]{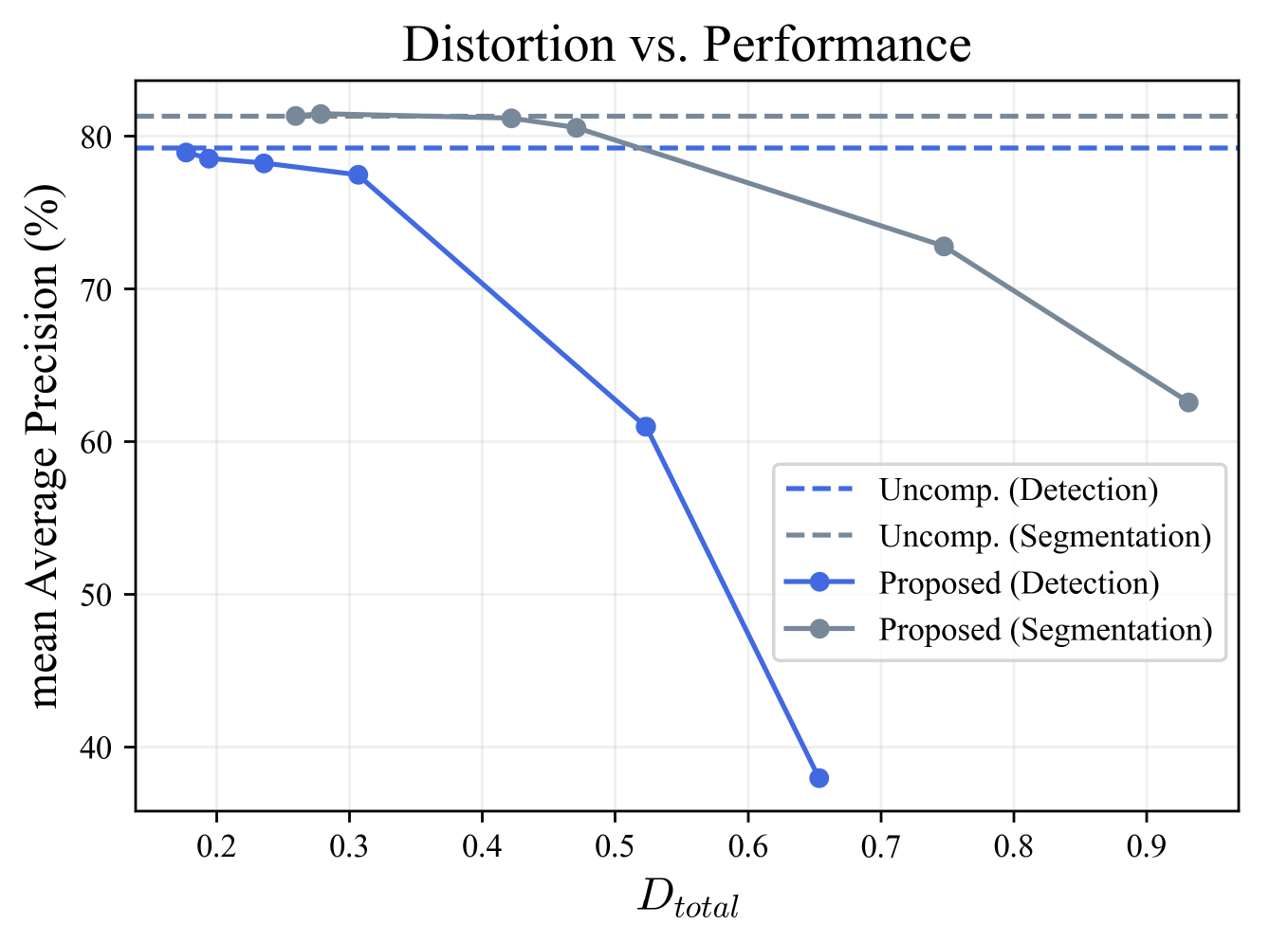}
\caption{Effects of distortion on the task performance of the proposed model.}
\label{fig12}
\end{figure}

\subsubsection{Compression Performance}
Table \ref{table:BD-rate gains} shows BD-rate reductions achieved by our model against the anchor models \cite{feature-anchor, image-anchor} and the previous methods \cite{kim2022, han2022, zhang2022}. In object detection task, the proposed model achieved 93.95\% and 98.22\% BD-rate reductions against the image anchor and feature anchor, respectively. It outperforms \cite{kim2022} with 52.23\% BD-rate reduction. In instance segmentation task, the proposed model achieved 95.79\% and 98.85\% BD-rate reductions against the image anchor and feature anchor, respectively. It outperforms \cite{zhang2022} with 84.58\% BD-rate reduction. We also evaluate the proposed model without the context model. In object detection task, {the proposed model (w/o CM)} achieved 91.06\% and 97.38\% BD-rate reductions against the image anchor and feature anchor, respectively. It outperforms \cite{kim2022} with 28.96\% BD-rate reduction. In instance segmentation task, {the proposed model (w/o CM)} achieved 85.1\% and 97.21\% BD-rate reductions against the image anchor and feature anchor, respectively. It outperforms \cite{zhang2022} with 60.47\% BD-rate reduction. Table \ref{table:rate-points} shows the actual values of bpp, $D_{total}$, and mAP for each quality level of the proposed model, along with the uncompressed feature data size in bpp.

Experimental results demonstrate that our L-MSFC approach attains superior compression performance compared to S-MSFC models \cite{kim2022, han2022}. This is attributed to the integration of multi-scale feature fusion with a learnable compressor, which enables end-to-end optimization for the multi-scale feature. Notably, our approach achieves these results without any task-specific loss. Our approach also outperforms the existing learned feature compression approach \cite{zhang2022}. Since we have used the same entropy model as \cite{zhang2022}, the performance improvement is solely attributed to the proposed FENet and DRNet. {Although $y$, the output of FENet, has a lower spatial dimension than each of the multi-scale feature maps, DRNet could successfully reconstruct all the feature maps with significantly higher task performance than \cite{zhang2022} at similar bitrate.} This also supports our previous claim that the proposed model can effectively reduce redundancy among layers.


\subsubsection{Near-Lossless Performance}
Table \ref{table:crnl-points} shows $R_{NL}$ (in bpp) and $CR_{NL}$ of our models and reference models. According to this table, the proposed models (with or w/o CM) require the smallest bitrate to reach the near-lossless mAP values. In other words, our models showed the highest compression ratio, $CR_{NL}$, at the near-lossless condition. Compared with the uncompressed case, the proposed model (with CM) achieved near-lossless performance with 27,555 times less bitrate for object detection task and 44,426 times less for instance segmentation task. Additionally, the proposed model (w/o CM) achieved near-lossless performance with 31,505 times less bitrate for object detection task and 38,680 times less for instance segmentation task.

\begin{figure*}[!t]
\centering
\includegraphics[width=\linewidth]{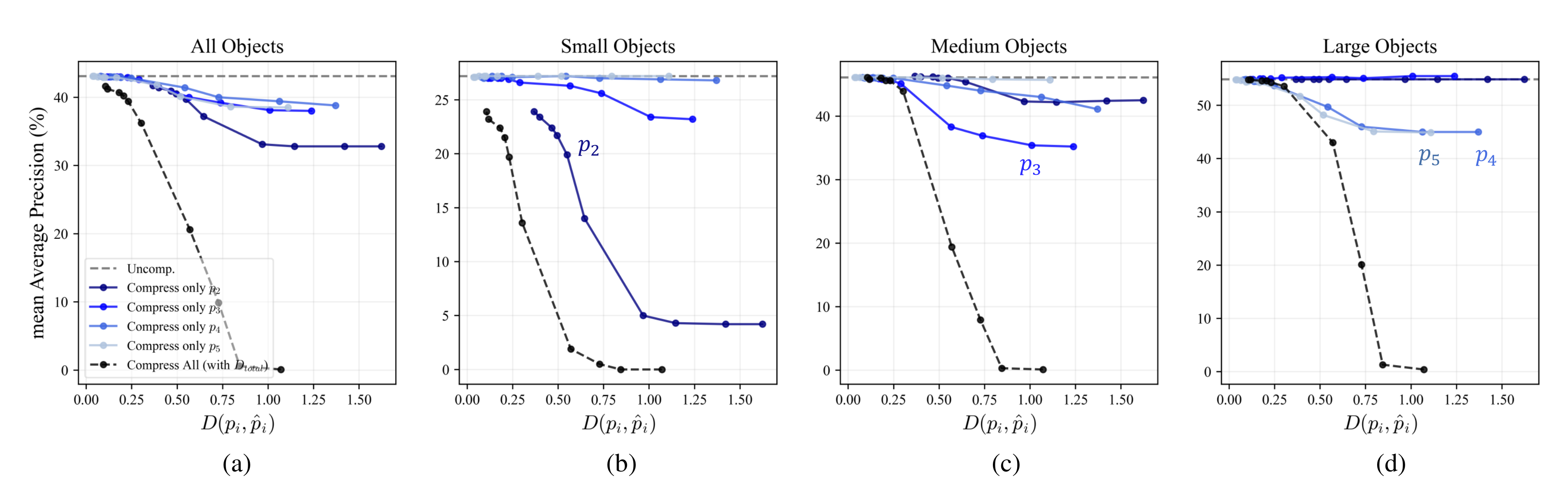}
\caption{{Distortion vs. mAP curves, where each mAP curve is generated by compressing only one layer of feature maps. (a) shows the mAPs for all objects, while (b), (c), and (d) measure the mAPs for different object sizes (i.e., small, medium, and large).}}
\label{fig13}
\end{figure*}

\begin{table}[t!]
    \renewcommand{\arraystretch}{1.2}
    \centering
    \caption{Summary of experimental results for the layer-wise distortion analysis}
    \resizebox{\linewidth}{!}{
    \begin{tabular}{l ccc cccc}
    \toprule
    & & & & \multicolumn{4}{c}{Layer-wise Distortion} \\
    \cmidrule(lr){5-8}
    $\lambda$ (quality level) & bpp & $D_{total}$ & mAP & $p_{2}$ & $p_{3}$ & $p_{4}$ & $p_{5}$ \\
    \midrule
    $\lambda_{1}=0.0025$    &0.0015&	1.0709&	0.1& 1.6234&	1.2389&	1.3705&	1.1092 \\
    $\lambda_{2}=0.0050$    &0.0025&	0.8441& 0.6&	1.4209&	1.0091&	1.0628&	0.7945 \\
    $\lambda_{3}=0.0125$ &  0.0054&   0.7275&   9.9&	 1.1453&	  0.7384&	0.7280&	0.5179 \\
    $\lambda_{4}=0.025$  &  0.0096&   0.5697&	20.6&    0.9674&	  0.5655&	0.5421&	0.3894 \\
    $\lambda_{5}=0.125$ &   0.0392&   0.3022&	36.2&    0.6452&	  0.2896&	0.2478&	0.1665 \\
    $\lambda_{6}=0.25$  &   0.0715&   0.2306&	39.4&    0.5490&	  0.2279&	0.1891&	0.0938 \\
    $\lambda_{7}=0.375$ &   0.0969&   0.2063&	40.2&    0.4944&	  0.1870&	0.1532&	0.1032 \\
    $\lambda_{8}=0.5$   &   0.1193&   0.1801&	40.7&    0.4656&	  0.1690&	0.1370&	0.0648 \\
    $\lambda_{9}=0.75$   & 0.1437&	0.1176&	41.2& 0.3987&	0.1302&	0.0912&	0.0428 \\
    $\lambda_{10}=1.0$   &0.1793&	0.1058&	41.6& 0.3673&	0.1158&	0.0792&	0.0363 \\
    \midrule
    Uncompressed&   680.0 & - & 43.1 & - & - & - & - \\   
    \bottomrule
    \end{tabular}
    }

    \label{table:summary of layer-wise analysis}
\end{table}

\subsubsection{Complexity}
In this subsection, the encoding and decoding time of the proposed methods and \cite{kim2022} are compared. We did not compare with \cite{zhang2022} since it will have a much large encoding and decoding time than ours due to its layer-wise repeated coding scheme. For the experiment, we used an NVIDIA GeForce RTX 3090 GPU and an AMD EPYC 7313 CPU. Fig. \ref{fig10}(a) shows that both of our models have higher compression performance than \cite{kim2022} with much less inference time. The inference times are measured excluding the feature extractor and task networks and averaged over different quality levels. Fig. \ref{fig10}(b)-(c) shows the encoding and decoding time over three different quality levels of each model. From Fig. \ref{fig10}(b), we can see that the proposed models have much less encoding time compared with \cite{kim2022} regardless of whether the context model is used or not. The higher the quality level, the longer encoding time \cite{kim2022} takes whereas our model maintains nearly constant encoding complexity. Note that a lightweight encoder design is very desirable in many VCM applications, especially when the features are encoded at low-complexity edge devices, e.g. CCTVs, and collected at servers or clouds for further analysis. In terms of decoding time, our model without the context model is comparable to \cite{kim2022}.

\subsection{Additional Analyses}
\subsubsection{{Feature Mixing Pathways in DRNet}}

In the proposed DRNet, the information in the lower-layer feature maps is propagated to upper-layer ones via the bottom-up pathway. However, the mixing pathway can be inverted so that the upper-layer features are propagated down to lower-layer ones via the top-down pathway. The top-down feature mixing blocks are implemented by replacing convolution layers in the bottom-up ones, shown in Fig. \ref{fig7}(d), with transposed convolution layers and switching $\hat{p}_{H}$ with $\hat{p}_{L}$.

{As an ablation, we compare the performances of the two pathways with the proposed model (w/o CM). Fig. \ref{fig11} shows that the bottom-up pathway outperforms the top-down one, obtaining a 17.50\% BD-rate reduction. Based on this result, we can conclude that the bottom-up pathway is the better choice for our model. Note that the top-down approach may be more efficient for parallel processing of the layer-wise branches.}


\subsubsection{Relationship between Distortion and Task Performance}
Fig. \ref{fig12} shows a plot of ($D_{total}$, mAPs) pairs from Table \ref{table:rate-points}. As shown in the figure, starting with uncompressed, mAP {monotonically decreases} as distortion increases, proving that the proposed rate-distortion optimization operates quite well.

\subsubsection{{Layer-wise Distortion Analysis}}
We designed an experiment to analyze the impact of distortions in each feature map layer on task performance. {We set the uncompressed task performance as a baseline and measured the individual performance degradation caused by the distortion of each feature map. More specifically, to measure the mAP score for the $i^{th}$ layer, we first compressed all the feature maps $\{p_{2}, p_{3}, p_{4}, p_{5}\}$ together with our model, reconstructed $\{\hat{p}_{2}, \hat{p}_{3}, \hat{p}_{4}, \hat{p}_{5}\}$, and replaced the uncompressed feature map $p_i$ with the reconstructed one $\hat{p}_i$, and then used the resulting feature maps $(\{p_{2}, p_{3}, p_{4}, p_{5}, p_{6}\}\setminus\{p_{i}\})\cup\{\hat{p}_{i}\}$ for inference through the task network.} The experiment was conducted on object detection task, using the proposed model (w/o CM). {For better analysis, we trained the model with additional lambda values (two lower ones and two higher ones), resulting in ten quality levels.} 5,000 images from the MS-COCO \cite{lin2014microsoft} validation set were used in this experiment. This dataset contains objects of various sizes, ranging from small to large. We used IoU thresholds of 0.5 to 0.95 in this experiment unlike other experiments conducted in this paper.

{Table \ref{table:summary of layer-wise analysis} and Fig. \ref{fig13} summarize the result of this layer-wise distortion experiment. In these table and figure, the distortion in $\hat{p}_{6}$ is not shown because $\hat{p}_{6}$ is simply down-sampled from $\hat{p}_{5}$, while $D_{total}$ represents averaged distortion of all the compressed feature maps (i.e., from $\hat{p}_{2}$  to $\hat{p}_{6}$). As shown in Fig. \ref{fig13}(a), although each individual feature map has a similar level of impact on task performance, $p_{2}$ had the most substantial influence on the overall performance.} However, when performance is measured over three different object sizes, the degradation is dominated by only one or two layers of feature maps. For example, in Fig. \ref{fig13}(b), for small objects, the highest-resolution feature map $p_2$ has the dominant influence, while the impacts of $p_4$ and $p_5$ are negligible. On the other hand, in Fig. \ref{fig13}(d), for large objects, the two lowest-resolution feature maps (i.e., $p_4$ and $p_5$) have a large influence on task performance, while the other layers show no impact.

\section{Conclusion}\label{sec:conclusion}
We proposed an end-to-end trainable multi-scale feature compression model which integrates feature fusion and a learnable compressor. Our model is evaluated under MPEG-VCM common test condition for object detection and segmentation tasks. The results showed that our model provides not only the highest compression performance but also the best complexity-performance trade-off with much less encoding time compared with the existing methods. Furthermore, our model achieves near-lossless {task performance} with at least 20,000 times less bitrate than the uncompressed one. In the future, we will extend the proposed model to support video tasks such as object tracking, where temporal redundancy between sequential frames should be importantly considered.

\bibliographystyle{IEEEtran}
\bibliography{ref}

 \begin{IEEEbiography}[{\includegraphics[width=1in,height=1.25in,clip,keepaspectratio]{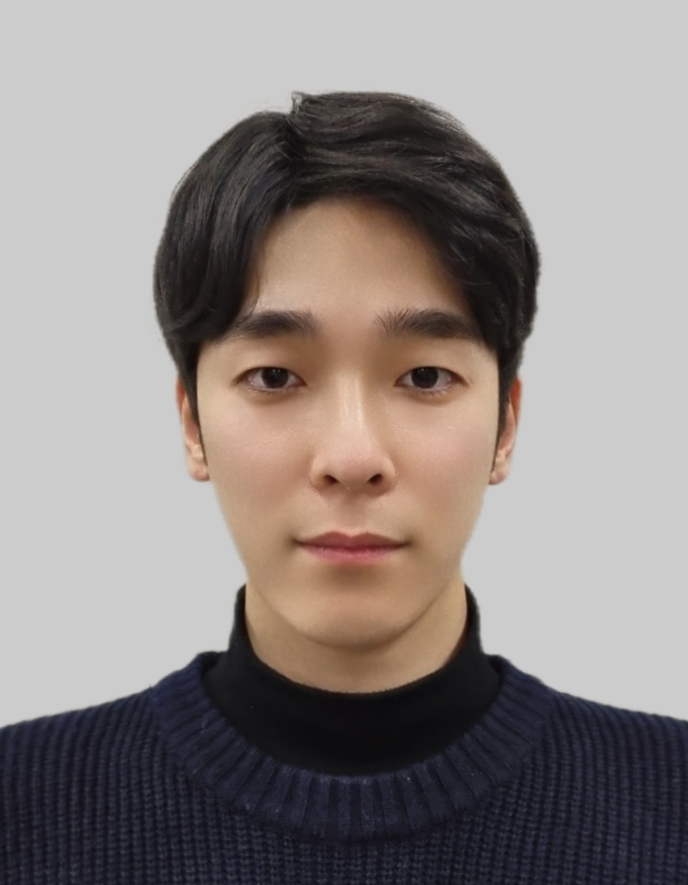}}]{Yeongwoong Kim} received his bachelor’s degree from the Department of Software Convergence and Department of Applied Sciences, Kyung Hee University, Yongin, South Korea, in 2022. Currently, he is pursuing an M.S. degree from the Department of Computer Science and Engineering at Kyung Hee University, South Korea. His research interests include video coding using deep neural network, computer vision and international standardization for video coding. 
\end{IEEEbiography}

\begin{IEEEbiography}[{\includegraphics[width=1in,height=1.25in,clip,keepaspectratio]{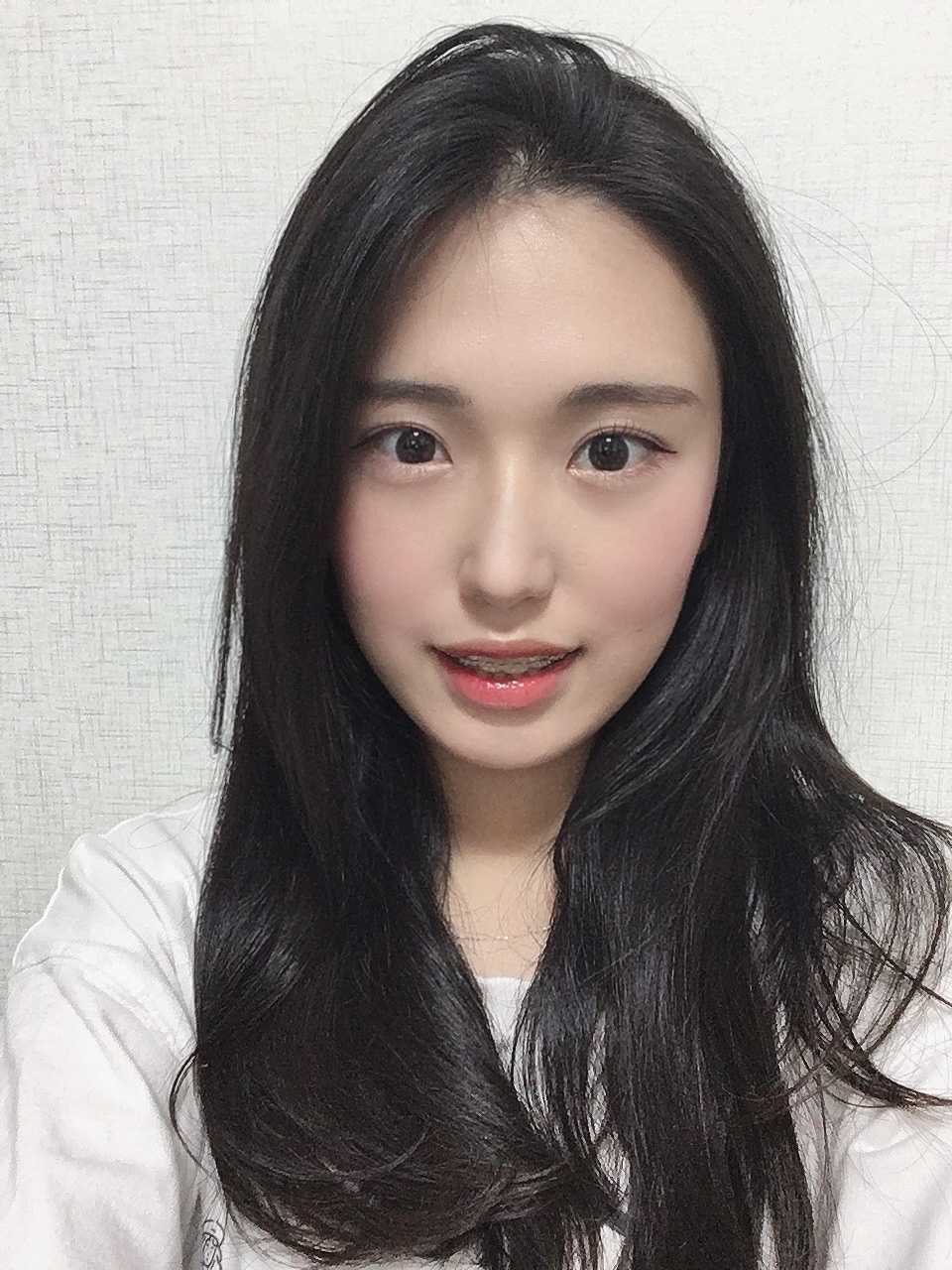}}]{Hyewon Jeong} received her bachelor’s degree from the Department of Software Convergence, Kyung Hee University, Yongin, South Korea, in 2022. Currently, she is pursuing an M.S. degree from the Department of Computer Science and Engineering at Kyung Hee University, South Korea. Her research interests include image processing, video coding, computer vision, machine learning, and international standardization for image and video coding. 
\end{IEEEbiography}

\begin{IEEEbiography}[{\includegraphics[width=1in,height=1.25in,clip,keepaspectratio]{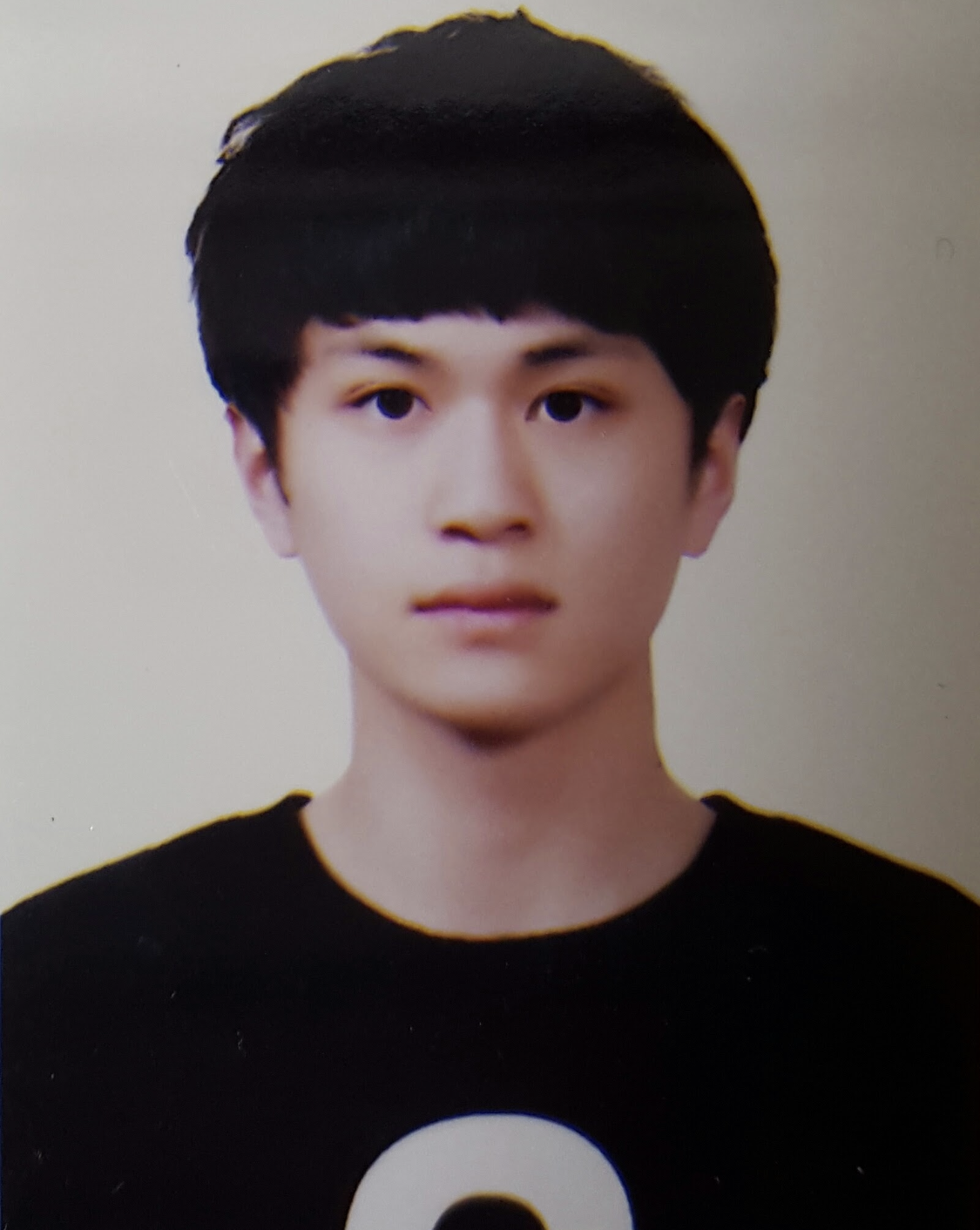}}]{Janghyun Yu} received his bachelor's degree from the Department of Electronic Engineering and Applied Sciences, Kyung Hee University, Yongin, South Korea, in 2023. Currently, he is pursuing an M.S. degree from the Department of Computer Science and Engineering at Kyung Hee University, South Korea. His research interests include video coding, computer vision, machine learning, and international standardization for video coding.
\end{IEEEbiography}

\begin{IEEEbiography}[{\includegraphics[width=1in,height=1.25in,clip,keepaspectratio]{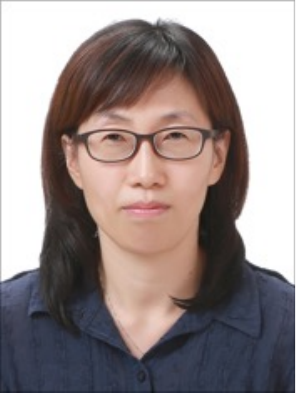}}] {Younhee Kim} received her BS and MS in computer science from Ajou University, Suwon, Rep. of Korea, in 2000 and in 2002, respectively, and her PhD in computer science from George Mason University, Fairfax, VA, USA, in 2009. She has been a senior researcher with ETRI, Daejeon, Rep. of Korea, since 2009. Her current research interests include video coding, image and video signal processing, and computer vision.
\end{IEEEbiography} 

\begin{IEEEbiography}[{\includegraphics[width=1in,height=1.25in,clip,keepaspectratio]{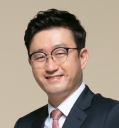}}] {Jooyoung Lee} received the B.S. degree in multimedia from Ajou University, Suwon, South Korea, in 2003, and the M.S. degree in computer science from Korea Advanced Institute of Science and Technology (KAIST), Daejeon, Korea in 2006. He is currently pursuing the Ph.D. degree in electrical engineering from Korea Advanced Institute of Science and Technology (KAIST). Since 2006, he has worked for Electronics and Telecommunications Research Institute, Daejeon, Korea, as a Principal Researcher. He had been involved in standardization activities of 3D broadcasting systems in Advanced Television Systems Committee (ATSC). His current research interests include learned image and video compression, deep learning for image restoration, visual quality enhancement, and perceptual video coding.
\end{IEEEbiography} 

\begin{IEEEbiography}[{\includegraphics[width=1in,height=1.25in,clip,keepaspectratio]{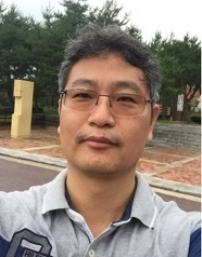}}]{Se Yoon Jeong} received the B.S and M.S degree from Inha University, in 1995, and 1997, respectively, and the Ph.D. degree from Korea Advanced Institute of Science and Technology, in 2014. In 1996, he joined Electronics and Telecommunications Research Institute, as a Principal Researcher. He has many contributions to development of international standards such as scalable video coding, high efficient video coding. His current research interests are in video coding for machine and AI based video coding.
\end{IEEEbiography} 

\begin{IEEEbiography}[{\includegraphics[width=1in,height=1.25in,clip,keepaspectratio]{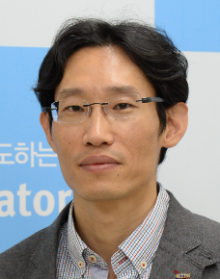}}]{Hui Yong Kim} received his BS, MS, and Ph.D degrees from KAIST in 1994, 1998, and 2004, respectively. He is currently an Associate Professor in the School of Computing in Kyung Hee University, Yongin, Korea. Prior to joining the university, he served as the Managing Director of Realistic Audio \& Video Research Group of Electronics and Telecommunications Research Institute (ETRI), Daejeon, Korea from 2005 to 2019. From 2003 to 2005, he worked for AddPac Technology Co. Ltd. as the Multimedia Research Team Leader. He has published more than 50 papers in international journals and conferences, and holds more than 400 registered patents over the world. He has been an active technology contributor, editor, and ad-hoc group chair in developing several international standards including MPEG Multimedia Application Format (MAF), ITU-T/ISO/IEC JCT-VC High Efficiency Video Coding (HEVC) and JVET Versatile Video Coding (VVC). His current research focus is neural-network based visual data compression, including learned video compression, digital hologram compression and deep feature compression for machine vision.
\end{IEEEbiography}

\end{document}